# Modular Pipe Climber

Rama Vadapalli
Robotics Research Center, IIIT Hyderabad
rama.raju@research.iiit.ac.in

Kartik Suryavanshi
Robotics Research Center, IIIT Hyderabad
suryavanshikartik@gmail.com

Ruchita Vucha
Robotics Research Center, IIIT Hyderabad
ruchita.vucha@students.iiit.ac.in

Abhishek Sarkar
Robotics Research Center, IIIT Hyderabad
abhishek.sarkar@iiit.ac.in

K Madhava Krishna
Robotics Research Center, IIIT Hyderabad
mkrishna@iiit.ac.in

## ABSTRACT

This paper discusses the design and implementation of the Modular Pipe Climber inside ASTM D1785 - 15e1 standard pipes [1]. The robot has three tracks which operate independently and are mounted on three modules which are oriented at 120° to each other. The tracks provide for greater surface traction compared to wheels [2]. The tracks are pushed onto the inner wall of the pipe by passive springs which help in maintaining the contact with the pipe during vertical climb and while turning in bends. The modules have the provision to compress asymmetrically, which helps the robot to take turns in bends in all directions. The motor torque required by the robot and the desired spring stiffness are calculated at quasi-static and static equilibriums when the pipe climber is in a vertical climb. The springs were further simulated and analyzed in ADAMS MSC. The prototype built based on these obtained values was experimented on, in complex pipe networks. Differential speed is employed when turning in bends to improve the efficiency and reduce the stresses experienced by the robot.

## 1. INTRODUCTION

Pipelines have been in use for transfer of oil, gases and fluids for a long time and recently small lightweight goods have also been transferred through pipelines by means of suction, known as pneumatic tube transport [3; 4]. Pipelines are generally installed underneath the surface to conceal their presence and to protect the pipe network from external damages [5]. Despite being ubiquitous, pipelines frequently fail due to corrosion and scaling by chemicals or by clogging. They are expensive to inspect, and it is often difficult to determine the exact location of the fault through external examination. In such situations, robotic inspection, where the robot travels inside the pipe and inspects for any cracks or faults by various Non-Destructive Methods (NDT) is feasible and promising [6].

Various robotic designs have been explored by researchers which include wheeled, caterpillar, articulated, inchworm, screw, Pipe Inspection Gauge (PIG) etc [7]. There has been extensive research on the multi-link wheeled robots, one of which Pipe Inspection Robot for Autonomous Exploration (PIRATE) was proposed by Dertien. et. al [8]. PIRATE is a robot for in-pipe inspection, which has omnidirectional wheeled modules connected in a zigzag manner to clamp against the inner wall of the pipe. The clamping forces of PIRATE can be controlled by a tensioning wire that runs along its body. PIRATE can also manipulate its motion with differential speed at the wheels. Although the design is robust, controlling the robot is hard because of the many active components involved. Research also exists in robotic designs involving a simple screw mechanism, with one motor for translation and another to adjust the direction of the screw. The design enables the robot to move helically inside the pipelines. Hirose designed the Thes-II robot, which could travel long distances in 50 mm inner-diameter pipelines with minimal energy consumption owing to its lesser number of motors [9]. However, the speed differential used in the robot makes its motion unpredictable in turns and fails to realize the intended motion inside the pipe. One of the most successful robots, the Multifunctional Robot for in-pipe Inspection (MRINSPECT) has worked in a range of pipeline diameters [10]. The 120° orientation of its driving modules makes the robot design robust and the passive parallelogram clamping mechanism of the modules works well in a range of pipe diameters. MRINSPECT has a robust design, but the links connecting the wheels, make contact with the pipe while taking turns and thereby increases the stress on the modules of the robot. Also, simple fluid driven pipe inspection robots has been used for cleaning, de-clogging, and inspection [11]. However, such robots follow the determined path of the pipeline and cannot be deployed in complex pipe networks [12]. Also, the robots with wheels tend to lack traction on some surfaces and experience wheel sinking problems [13].

We have designed the Modular Pipe Climber (Fig. 1) keeping in mind the industrial factors which required the robot to be simple and robust in its working. The design intent for the Modular Pipe Climber was to build a robot that is simpler in design with less moving parts, which is easier to build and implement compared to the currently existing pipe climbers. The Modular Pipe Climber has tracks instead of wheels, which provide a greater surface for contact and in return greater traction [2]. The robot has three modules which are separated by 120° and connected with the center chassis by four shafts each, with each shaft equipped with a spring which keeps the robot pressed against the inner wall of a pipe during vertical climb and when turning in bends.

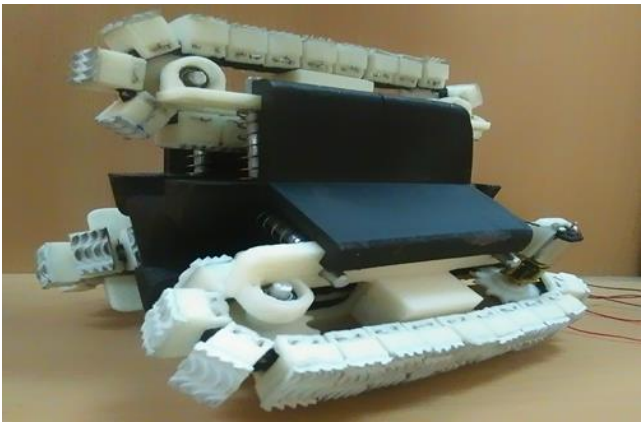

**Figure 1: Working Prototype of the Modular Pipe Climber**

The design and dynamics of the Modular Pipe Climber are presented in Section 2. In Section 3, design optimization, calculations are made to determine the required spring stiffness, torque and the dimensional constraints. In Section 4, in-pipe locomotion, motion study is conducted on the Modular Pipe Climber in MSC ADAMS to analyze it vertical climbing and turning capabilities and experiments were conducted on the prototype of the robot in vertical pipes and bends.

## 2. DESIGN & DYNAMICS

### A. Design Intent:

The Modular Pipe Climber is designed to climb vertically and turn in 30º, 45º, 60º and 90º bends inside Polyvinyl Chloride (PVC) pipes of the standard "ASTM D1785 - 15e1" with Schedule (Sch) 40, 80 and 120 and Nominal Pipe Size (NPS) 6[1].

As the effective diameter of the pipe continuously changes during bends, to turn in bends, the robot is required to vary its diameter to maintain contact with the inner wall of the pipe [14]. The Modular Pipe Climber maintains contact with the inner wall of the pipe with the help of its pre-loaded springs, which enables it to manipulate its diameter from 163.33 mm to 129.54 mm (Fig. 2).

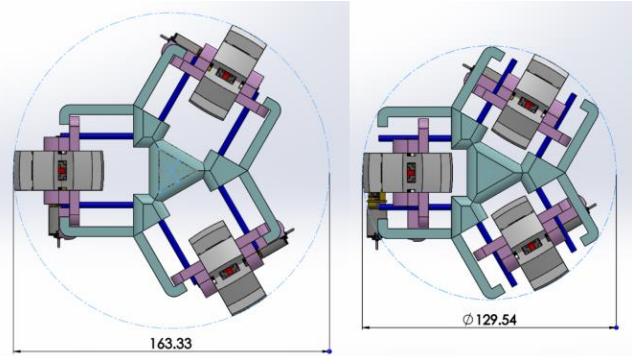

**Figure 2: Maximum and Minimum Diameters of the Modular Pipe Climber**

To keep the changes in the diameter of the robot when turning in bends as low as possible while maintaining traction, the length of the robot needs to be close to the value of its diameter. The Modular Pipe Climber is designed with a length of 150 mm, which is in-between its maximum and minimum diameters as discussed in Section 3C (Dimensional Constraints).

### B. Components overview:

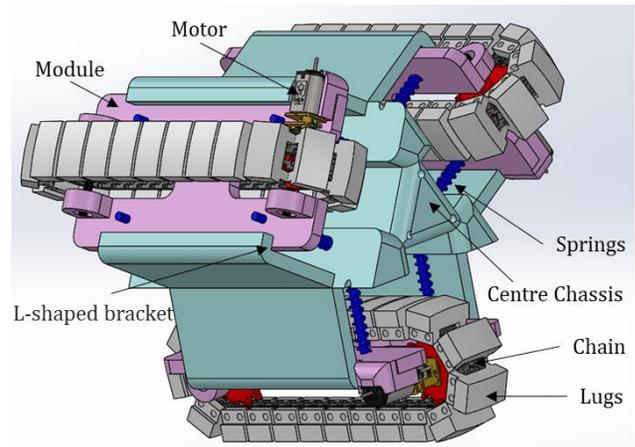

**Figure 3: CAD Model of the Modular Pipe Climber**

The Modular Pipe Climber has three tracks which operate independently on three modules and are oriented at 120° to each other (Fig. 3). Each track is powered by a separate 1000:1 Micro Metal Gearmotor HPCB 12V Pololu Gearmotor which generate 10 kg-cm Torque at 35 RPM [15]. The modules are assembled around the center chassis and are pre-loaded with linear springs which push the tracks radially outwards to provide traction against the inner wall of the pipe. The linear springs also allow the Modular Pipe

Climber to vary its diameter, which enables it to turn in bends and to be used in pipes of varying diameter. The springs are placed around shafts that are attached to the center chassis and protrude out of the modules. The shafts ensure that the springs remain in place and the module moves along its path. The center chassis' L-shaped bracket (Fig. 3) ensures that the springs do not push the modules out of place and the Modular Pipe Climber's diameter does not exceed 163.33 mm. Each track is assembled with 22 lugs of which at least 9 maintain contact with the pipe during the vertical climb. The lugs have a curved outer surface with a radius of curvature of 80 mm, which is approximately the same as the radius of the pipes chosen for the robot's design. This enables the lugs to establish maximum contact with the cylindrical inner wall of the pipe. Each lug is equipped with a Latex layer on its curved outer surface to increase the friction between the tracks and the inner wall of the pipe.

## 3. DESIGN OPTIMIZATION

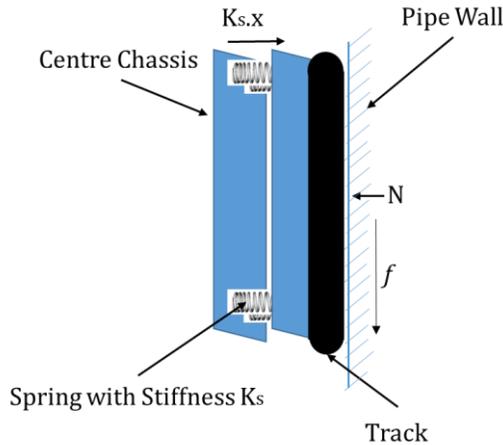

**Figure 4: Schematic Diagram of a Module in the Pipe**

### A. Calculation of Spring Stiffness:

The optimum spring stiffness is calculated at static equilibrium when the robot is in a vertical climb. Balancing the forces for a single module in the horizontal and vertical directions and assuming the normal force ($N$) is distributed equally amongst the four pre-loaded springs, we get

$$4K_s x = \frac{N}{3}, \tag{1}$$

and when equating the forces acting on the pipe climber along the vertical direction, we get

$$\mu_s N = mg, \tag{2}$$

where $K_s$ is the spring stiffness, $x$ is the change in length of the spring, $\mu_s$ is the coefficient of static friction, $m$ is mass of the robot and g is the acceleration due to gravity. Combining the above two equations we get

$$12\mu_s K_s x = mg. \tag{3}$$

### B. Calculation of Torque:

Motor torque is calculated at quasi-static equilibrium during the robot's vertical climb. Since the lugs themselves have almost no compression, it is assumed that the rolling resistance ($RR$) is negligible.

$$RR = N.C_R \approx 0, \tag{4}$$

where $C_R$ is the coefficient of rolling resistance. The friction ($f$) experienced by the robot during the vertical climb is calculated taking into consideration only the sliding friction since it is assumed that there is no rolling resistance.

$$f = 12\mu_k K_s x, \tag{5}$$

where $\mu_k$ is the coefficient of sliding friction. The inertial force ($F_a$) of the robot during the vertical climb

$$F_a = ma, \tag{6}$$

where $a$ is the acceleration of the robot inside the pipe. The total tractive effort ($TTE$) of the robot during the vertical climb is

$$TTE = RR + F_a - f + mg, \tag{7}$$

By substituting rolling resistance ($RR$) from equation (4), inertial force ($F_a$) from equation (5) and friction ($f$) from equation (6) in equation (7), we get

$$TTE = ma + mg - 12\mu_k K_s x, \tag{8}$$

where $\mu_k$ is the coefficient of sliding friction. The required torque ($\tau$) of the motor

$$\tau = TTE.r_{wheel}, \tag{9}$$

where $r_{wheel}$ is the radius of the wheel, which is half the height of the track.

By substituting the values of mass of the robot $m$ = 470 g, the change in length of the spring $x$ = 26 mm and coefficient of sliding friction $\mu_k$ = 0.7 in equations (3) and (9) we get spring stiffness, $k_s$ = 18.06 N/m, and motor torque, $\tau$ = 0.23 N-m.

To keep a high factor of safety and to account for various other unaccounted losses, motors with 0.88 N-m of torque were used.

## C. Dimensional Constraints

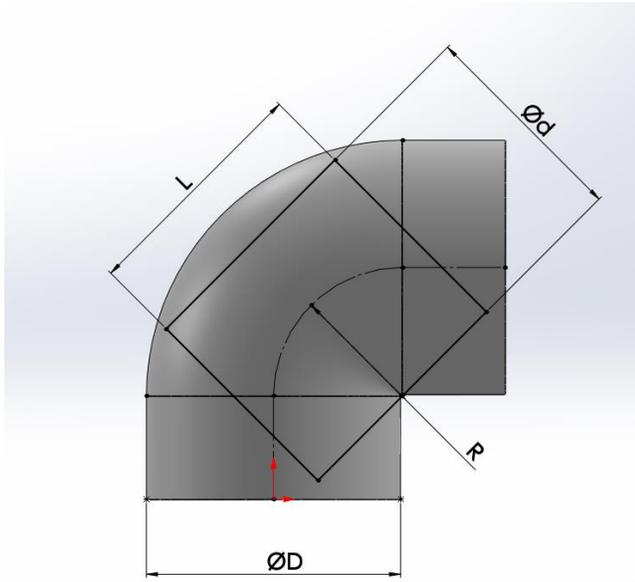

**Figure 5: Simplified Two-Dimensional Adaption of Modular Pipe Climber in a 90º Bend**

The robot is designed such that its diameter is comparable to its height. To calculate the dimensional relation between the Modular Pipe Climber and the 90º bend, a simplified two-dimensional view of the Modular Pipe Climber (Fig. 5) is taken.

For the robot to be able to negotiate the 90º bend the following equation needs to be satisfied [16].

$$(R + D/2)\sin 45^0 - (R - D/2) < d < D, \quad (10)$$

where $R$ is the radius of curvature of the pipe bend, $D$ is the diameter of the pipe and d is the minimum diameter of the robot required to turn in a 90º bend.

By substituting the values $R = 90$ and $D = 160$ mm in equation (10), we find that the minimum diameter of the robot to go through the 90º bend should be between 110-160 mm. The minimum diameter d chosen for the Modular Pipe Climber is 129.54 mm.

The required length ($L$) of the robot to turn in a 90º bend

$$L = \sqrt{(R + D/2)^2 - (R - D/2 + d)^2}, \quad (11)$$

For the Modular Pipe Climber whose minimum diameter is $d$ = 129.54 mm, to turn in a 90º bend with an inner diameter of $D = 160$ mm and a radius of curvature of $R = 90$ mm, the length of the robot is 150 mm.

## 4. IN-PIPE LOCOMOTION

### A. Motion Study:

To validate the analytical values found in the above equations, a motion study was conducted on MSC ADAMS. Since the tracks were too complex to model in the software, a simplified lumped model (Fig. 6) was made and analyzed with spring stiffness ranging from 16 N/m to 26 N/m. The model was first tested for vertical climbing and later for 45º, 90º smooth bends. The results showed that the model was able to successfully climb vertically and turn in bends without any slip and with fewer frictional losses when the springs used had a spring stiffness of 18.06 N/m as calculated in Section 3B.

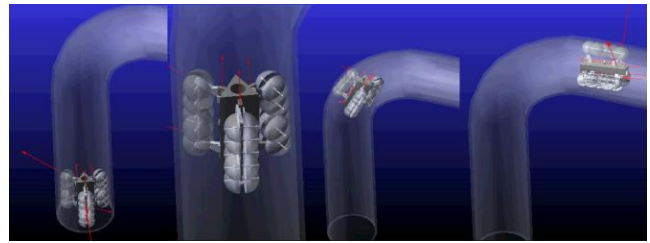

**Figure 6**: **Simplified Lumped Model of the Modular Pipe Climber Taking Turn in a 90º Smooth Bend**

To explain the navigation of the Modular Pipe Climber when taking a turn, the deformation of its springs is studied. In order to calculate the maximum deformation of the springs, the robot is aligned such that one of the 3 modules (outer module) follows the longest possible path inside the pipe when taking a turn and the other two modules (inner modules) follow a shorter path. The plot (Fig. 7) shows the spring compression results of the motion study in MSC ADAMS of the front and rear springs of the inner and outer modules of the robot when it moves through a 90º bend. The springs' compression at its pre-loaded state during the robot's vertical climb is taken as 0 in the plot. When the spring is further compressed from its pre-loaded state, it is reflected in the plot as a positive change in the spring's length and when the module is given freedom to expand beyond its pre-loaded state it reflects as a negative change in its length. The plot shows that there is a delay in response of the rear springs to that of the front springs when the robot moves through a 90º bend. This occurs because, when turning inside a bend at a constant velocity of 100mm/sec (Fig .7), the front of the pipe climber enters the turn first at approximately 0.6 seconds in the plot causing the front springs to compress and with a delay of approximately 0.3 seconds the rear of the module enters the bend and begins to compress the rear springs. The plot also shows that after going through the bend at approximately 1.4 seconds the front springs momentarily experience expansion from its initial pre-loaded state. This expansion happens because after passing through the bend the front of the robot temporarily has more freedom to expand.

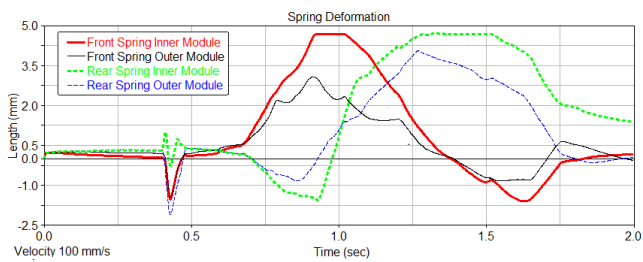

**Figure 7**: Spring Compression of Front and Rear Springs of the Inner and Outer Modules

The plot further indicates that the springs on the outer module undergo less compression compared to that of the springs on the inner module. This variation occurs because when turning in a $90^0$ bend the robot's orientation with respect to the ground changes from vertical to horizontal and majority of its weight falls on the springs of the inner module and causes additional compression to the spring.

**B. Experimentation:**

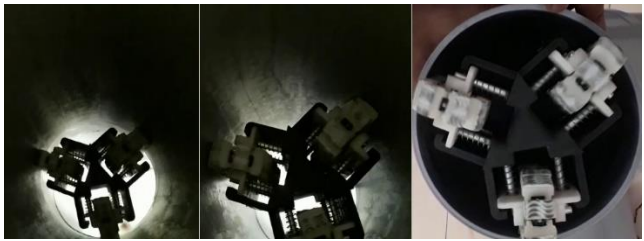

**Figure 8**: Modular Pipe Climber Climbing Vertically in a Pipe

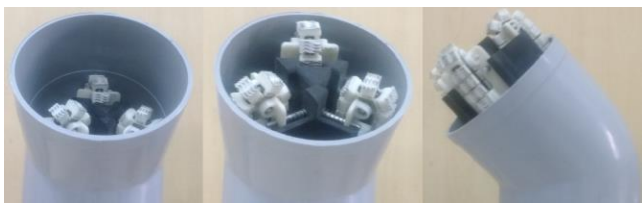

**Figure 9**: Modular Pipe Climber Negotiating a Turn in a 45º Bend

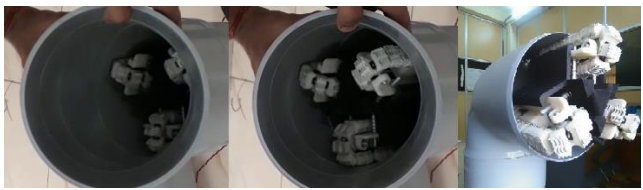

**Figure 10**: Modular Pipe Climber Negotiating a Turn in a 90º Bend

A prototype of the Modular Pipe Climber was built to the exact scale of its CAD design to experiment on and observe its vertical climbing capabilities and how it negotiated the bends. The center chassis, modules and the lugs were 3D printed with ABS [Acrylonitrile Butadiene Styrene] and were assembled together with the motors, tracks, springs and the shafts. Aluminium shafts with a circular cross-section of 4 mm diameter were used which allowed for a smoother sliding of the modules. The prototype was extensively tested in various pipe networks and was found to be successful in climbing pipes vertically (Fig .8) and taking turns in 45º (Fig .9) and 90º bends (Fig .10). Although the three tracks could operate independent to each other, during the initial testing, all three tracks were controlled through a single connection and were manually operated by a DPDT [Double-Pole, Double-Throw] switch. During the vertical climb, a constant velocity was maintained in all tracks which helped the robot to climb the pipe efficiently. While taking a turn in bends, it was observed that the inner track which had a lesser distance to travel tended to slip as all the tracks were driven at the same speed. To avoid slippage in bends, the speed of each track was adjusted according to the bend. The ratio of speeds of the outer module to the inner module was set at 25.4:10, which is equal to the ratio of the radius of curvature of the paths followed by the outer track to the inner track respectively in a 90º bend. This led to a much fluent motion and fewer stresses acting on the tracks.

## 5. CONCLUSION & FUTURE WORK

We have designed the Modular Pipe Climber keeping in mind the industrial factors which required the robot to be simple and robust in its working. In motion study and during experimentation, the Modular Pipe Climber successfully climbed vertically and negotiated 45º and 90º bends inside a variety of pipe networks. The robot has few moving parts and requires no complex motions to climb vertically or to turn in bends. This makes the robot robust and easy to be implemented in real-world applications. The differential speed used in the Modular Pipe Climber enhances the robot's capability to turn in bends with ease. Although the robot was made with the intent of keeping its design simple, the Modular Pipe Climber is able to negotiate bends of all angles between 0º and 90º. Furthermore, the Modular Pipe Climber is easy to operate and requires minimal maintenance. This makes the robot well suited for in-pipe surveillance and inspections in industries.

Presently the Modular Pipe Climber cannot negotiate junctions. Our future work involves designing a robot that can turn in complex junctions.

## REFERENCES

[1] Standard specification for Poly(Vinyl Chloride)(PVC) Plastic Pipe, shttps://www.astm.org/Standards/D1785.htm.
[2] J. Y. Wong and Wei Huang John. "Wheels vs. tracks" – A fundamental evaluation from the traction perspective. Journal of Terramechanics. 43.1 (2006) : 27-42.
[3] John L. Kennedy. Oil and gas pipeline fundamentals. Pennwell books, 1993.
[4] Wojciech Poznanski, Frances Smith, and Frank Bodley." Implementation of a pneumatic-tube system for transport of blood specimens". American journal of clinical pathology. 70.2 (1978): 291-295.
[5] M. Ahammed and R. E. Melchers. Reliability of Underground Pipelines Subject to Corrosion. Journal of Transportation Engineering. 120.6 (1994) : 12-18.

[6]  Kawaguchi, Yoshifumi, et al." Internal pipe inspection robot". Proceedings of 1995 IEEE International Conference on Robotics and Automation. Vol. 1. IEEE, 1995.
[7]  Roslin, Nur Shahida, et al. A review: hybrid locomotion of in pipe inspection robot. Procedia Engineering 41 (2012): 1456-1462.
[8]  Dertien, Edwin, et al. Design of a robot for in-pipe inspection using omnidirectional wheels and active stabilization. Robotics and Automation (ICRA), 2014 IEEE International Conference on. IEEE, 2014.
[9]  Hirose, Shigeo, et al. Design of in-pipe inspection vehicles for/spl phi/25,/spl phi/50,/spl phi/150 pipes. Robotics and Automation, 1999. Proceedings. 1999 IEEE International Conference on. Vol. 3. IEEE, 1999.
[10] Se-gon Roh, and Hyouk Ryeol Choi. "Differential-drive in-pipe robot for moving inside urban gas pipelines". IEEE Transactions on Robotics 21.1 (2005): 1-17.
[11] Zheng Hu, and Ernest Appleton. "Dynamic characteristics of a novel self-drive pipeline pig". IEEE Transactions on Robotics 21.5 (2005): 781-789.
[12] Zhu, X., Wang, W., Zhang, S. et al. Experimental Research on the Frictional Resistance of Fluid-Driven Pipeline Robot with Small Size in Gas Pipeline. Tribol Lett (2017) 65: 49.
[13] T. Hiroma, S. Wanjii, T. Kataoka and Y. Ota. Stress analysis using fem on stress distribution under a wheel considering friction with adhesion between a wheel and soil. Journal of Terramechanics. 34.4 (1997) : 225-233.
[14] Iszmir Nazmi Ismail, Adzly Anuar, Khairul Salleh Mohamed Sahari, Mohd Zafri Baharuddin, Muhammad Fairuz, Abd Jalal and Juniza Md Saad. "Development of in-pipe inspection robot: A review". IEEE Conference on. Vol. 2. IEEE, 2013.
[15] Pololu Micro metal Gearmotor HPCB 12V, https://www.pololu.com/product/3046.
[16] Choi, H. R., and S. M. Ryew. "Robotic system with active steering capability for internal inspection of urban gas pipelines." Mechatronics 12.5 (2002): 713-736.